\documentclass[siggraph]{acmart}
\acmSubmissionID{837}

\usepackage{booktabs} 
\usepackage{enumitem}
\usepackage{subcaption}
\usepackage{balance}

\usepackage[ruled]{algorithm2e} 

\SetAlFnt{\small}
\SetAlCapFnt{\small}
\SetAlCapNameFnt{\small}
\SetAlCapHSkip{0pt}

\usepackage{amsmath}
\DeclareMathOperator*{\argmax}{arg\,max}
\DeclareMathOperator*{\argmin}{arg\,min}

\acmJournal{TOG}

\copyrightyear{2023}
\acmYear{2023}
\setcopyright{acmlicensed}\acmConference[SA Conference Papers '23]{SIGGRAPH Asia 2023 Conference Papers}{December 12--15, 2023}{Sydney, NSW, Australia}
\acmBooktitle{SIGGRAPH Asia 2023 Conference Papers (SA Conference Papers '23), December 12--15, 2023, Sydney, NSW, Australia}
\acmPrice{15.00}
\acmDOI{10.1145/3610548.3618238}
\acmISBN{979-8-4007-0315-7/23/12}

\begin{document}
\title{SeamlessNeRF: Stitching Part NeRFs with Gradient Propagation}

\author{Bingchen Gong}
\authornote{Both authors contributed equally to this work.}
\email{gongbingchen@gmail.com}
\orcid{0000-0001-6459-6972}
\author{Yuehao Wang}
\authornotemark[1]
\email{yhwang@link.cuhk.edu.hk}
\orcid{0009-0003-3144-128X}
\affiliation{%
  \institution{The Chinese University of Hong Kong}
  \city{Hong Kong}
  \country{China}
}

\author{Xiaoguang Han}
\authornote{Corresponding authors: hanxiaoguang@cuhk.edu.cn; qidou@cuhk.edu.hk}
\email{hanxiaoguang@cuhk.edu.cn}
\orcid{0000-0003-0162-3296}
\affiliation{%
  \institution{The Chinese University of Hong Kong (Shenzhen)}
  \city{Shenzhen}
  \country{China}
}

\author{Qi Dou}
\authornotemark[2]
\email{qidou@cuhk.edu.hk}
\orcid{0000-0002-3416-9950}
\affiliation{%
  \institution{The Chinese University of Hong Kong}
  \city{Hong Kong}
  \country{China}
}

\begin{abstract}
Neural Radiance Fields (NeRFs) have emerged as promising digital mediums of 3D objects and scenes, sparking a surge in research to extend the editing capabilities in this domain. The task of seamless editing and merging of multiple NeRFs, resembling the ``Poisson blending'' in 2D image editing, remains a critical operation that is under-explored by existing work. To fill this gap, we propose SeamlessNeRF, a novel approach for seamless appearance blending of multiple NeRFs. In specific, we aim to optimize the appearance of a target radiance field in order to harmonize its merge with a source field. We propose a well-tailored optimization procedure for blending, which is constrained by 1) pinning the radiance color in the intersecting boundary area between the source and target fields and 2) maintaining the original gradient of the target. Extensive experiments validate that our approach can effectively propagate the source appearance from the boundary area to the entire target field through the gradients. To the best of our knowledge, SeamlessNeRF is the first work that introduces gradient-guided appearance editing to radiance fields, offering solutions for seamless stitching of 3D objects represented in NeRFs.
Our code and more results are available at \url{https://sites.google.com/view/seamlessnerf}.
\end{abstract}

%
%

\ccsdesc[500]{Computing methodologies~Reconstruction}
\ccsdesc[300]{Computing methodologies~Image-based rendering}
\ccsdesc[300]{Computing methodologies~Volumetric models}

%
%

\keywords{neural radiance fields, gradient propagation, seamless, composition, 3D editing}

\begin{teaserfigure}
    \centering
    \includegraphics[trim={0 7 0 10},clip,width=\textwidth]{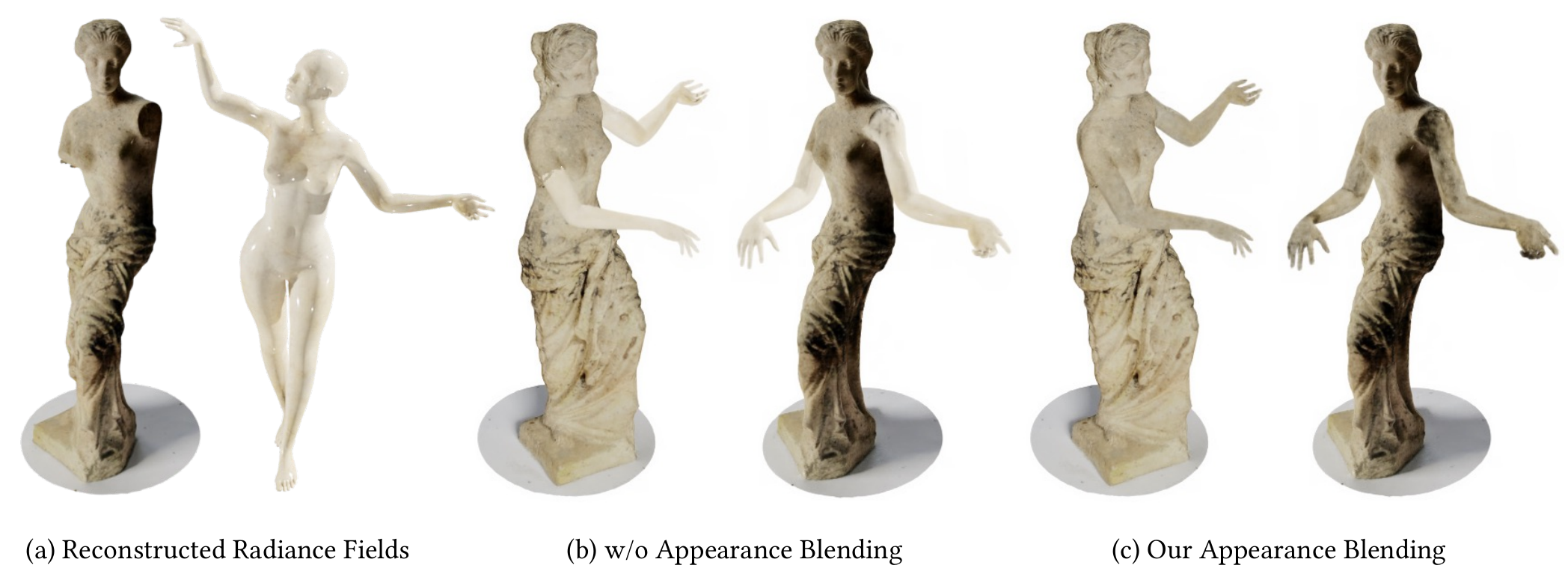}
    \caption{We demonstrate the appearance blending results of our proposed approach. In the exhibited case (a), we represent the sculptures in radiance fields and aim to repair the ``Venus de Milo'' sculpture with the arms of another marble statue. In (b), we display the results obtained by a direct concatenation of the target arms and the source sculpture, which incurs incoherence between the new arms and the original one. In (c), our proposed approach can overcome this problem by fusing the source appearance into the target parts.}
    \label{fig:teaser}
\end{teaserfigure}

\maketitle

\section{Introduction}

Neural Radiance Fields (NeRFs) \cite{mildenhall2021nerf} have been established as a potent representation of 3D scenes and are widely considered to be the top contender for future media forms, alongside images and videos. Therefore, incorporating basic editing functionalities into this novel representation is of substantial importance.


Seamless stitching and transplantation of a 3D object to another is a fundamental editing operation, akin to the common seamless blending functionality on 2D images. This is a long-standing problem \cite{rocchini1999multiple,yu2004mesh} as mixing and re-creating 3D models are important steps in gaming, filming, and artistic creation. For instance, as illustrated in Fig. 1, the arms of a ceramic dancing woman can be precisely extracted and seamlessly integrated into the renowned ``Venus de Milo'' sculpture for building artistic derivative. By this merging process, the texture attributes from the source (the ``Venus de Milo'') are propagated into the newly stitched arms, producing a natural and cohesive appearance. Nevertheless, the seamless editing technique is still missing for 3D objects represented in NeRFs. Since appearance information in radiance fields is encoded in black-box network parameters, propagating texture information between source and target implicit fields poses a particular challenge.


Among the various approaches developed to seamlessly merge two images, gradient-based melding techniques stand out due to their robustness to diverse scenarios and user-friendly interactions \cite{Darabi2012ImageM,Bhat2010GradientShopAG,Prez2003PoissonIE}. These methods, anchored on the Poisson equation, basically resolve a system of equations to uphold gradient consistency on the intersecting boundary of images and preserve the original image gradients within the boundary. In this way, a smooth transition is imposed between the merged images. In the domain of implicit fields, the manipulation of objectives involving gradients has been established as a viable pursuit, as affirmed by previous work \cite{xu2022signal,li2023neuralangelo}. Consequently, gradient-based melding emerges as an attractive option for facilitating the seamless merging of NeRFs.



In this paper, we present SeamlessNeRF, a novel framework designed to facilitate gradient-based appearance blending of NeRFs. Given source and target objects represented in NeRFs, our key idea is to optimize the appearance of the target field to harmonize its texture style with the source field, thereby achieving a smooth transition between the source and target fields. To do this, we begin with transforming the source and target NeRFs into a unified homogeneous coordinate space, followed by constructing a piecewise radiance field based on their density values.
During the optimization for appearance blending, the radiance of the source and target are reconciled in their intersecting boundary area. Concurrently, we impose regularization on the gradient field of the target to preserve its original appearance change. This enables the propagation of radiance color from the source to the target through the gradient field. In order to efficiently compute the radiance gradients in the unified 3D space, we resort to finite difference in implicit fields. Since radiance fields are view-dependent, we propose a sampling strategy to identify each point's view direction. Furthermore, a side-branch fine-tuning scheme is incorporated into the appearance optimization procedure for faster convergence and preserving texture details.
In summary, our main contributions include:
 \begin{itemize}[leftmargin=*]
	\item We propose a novel framework for seamlessly stitching part NeRFs into a unified and harmonized 3D representation, which, to our knowledge, is the first of its kind.
	\item We introduce a gradient propagation method, which preserves the texture patterns that are inherently implied in the gradient fields, and ensures a smooth radiance transition across the boundaries.
	\item We conduct comprehensive experiments to validate the effectiveness of our approach across a diverse range of scenarios and different backbones.
 \end{itemize}

\section{Related Work}
\label{sec:related}

\begin{figure*}[t]
	\centering
	\includegraphics[width=0.95\textwidth]{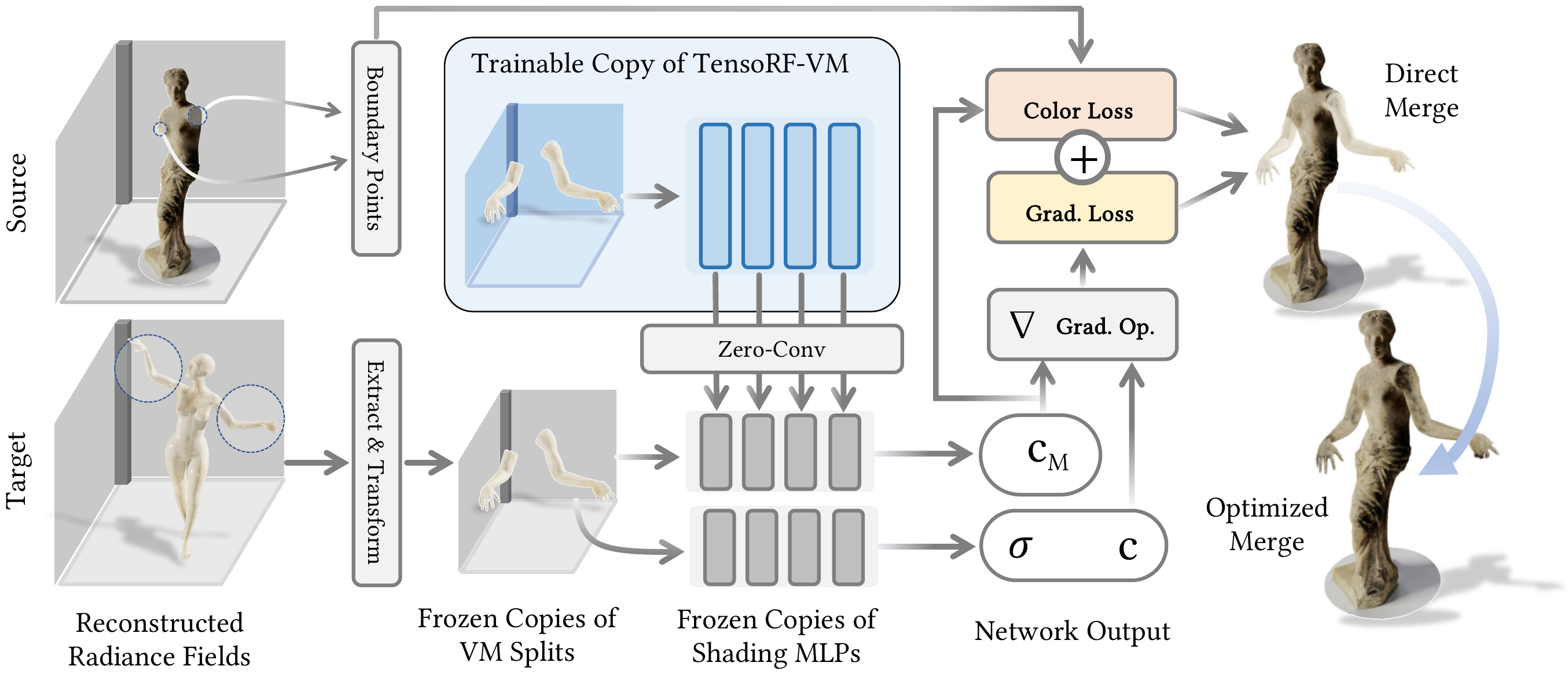}
	\caption{Pipeline of our proposed approach for seamless appearance blending. We adopt TensoRF-VM \cite{chen2022tensorf} as our radiance field backbone. The pipeline enables appearance blending by fine-tuning a  trainable copy of the target radiance field, which connects the frozen copies via zero convolutions.}
	\label{fig:pipeline-figure}
\end{figure*}

\subsection{NeRF Editing}

Previous work \cite{zhang2021editable,tang2022compressible} has explored possible solutions to entity placement and composition of NeRF-represented scenes. NeRFShop \cite{NeRFshop23} allows users to interactively select and deform objects through cage-based transformations, where membrane interpolation technique is proposed to reduce potential editing artifacts. Block-NeRF \cite{tancik2022block} merges multiple radiance fields into one in order to achieve novel view synthesis of large-scale scenes. To composite adjacent radiance fields under different time of day, color balance, and weather, an appearance matching scheme is proposed to align the appearance of the neighboring blocks. However, Block-NeRF considers merging two blocks of the same scene with large overlaps, which diverges from stitching part NeRFs.

Regarding appearance editing in NeRFs, the challenge arises from their implicit scene representations. Previous study \cite{yang2021learning, liu2021editing, zhuang2022mofanerf, zheng2022editablenerf, xu2022signal} focused on purely implicit editing. The approach proposed by \cite{yuan2022nerf} extracts explicit meshes as a manipulable proxy to facilitate editing in implicit representations.
NeRV \cite{srinivasan2021nerv}, NeRFactor \cite{zhang2021nerfactor}, SAMURAI \cite{boss2022samurai}, NVDiffRec \cite{munkberg2022extracting}, and Ref-NeRF \cite{verbin2022ref} support editable BRDF materials by decoupling 3D geometry and lighting effects from multi-view captures. Other noteworthy progresses in this area, including EditNeRF \cite{liu2021editing} and CodeNeRF \cite{yang2021learning}, encode the shape and appearance of simple objects into an interpolative latent code. CLIP-NeRF \cite{wang2022clip} and DFFs \cite{kobayashi2022distilledfeaturefields} explored text-guided scene editing  utilizing the joint vision-language embedding inherent in the CLIP model \cite{radford2021learning}. Instruct-NeRF2NeRF \cite{haque2023instruct} further enhances text-guided NeRF editing by using diffusion models \cite{brooks2023instructpix2pix}. PaletteNeRF \cite{kuang2023palettenerf} and RecolorNeRF \cite{gong2023recolornerf} enable palette-based color editing for NeRF-represented scenes. Style transferring schemes proposed by UPST-NeRF \cite{chen2022upst}, INS \cite{fan2022unified}, StyleRF \cite{liu2023stylerf}, and ARF \cite{zhang2022arf} stylize NeRF models to resemble reference images. NeRFEditor \cite{sun2022nerfeditor} goes a step further by encoding novel-view images into the hidden space of StyleGAN, supporting editing guided by reference images, text prompts, and user interactions.
%
%
%
%
While the majority of the methods discussed in this section have demonstrated significant capabilities in composition and appearance editing, they struggle when tasked with the seamless stitching and appearance blending of NeRF-represented assets.

\subsection{Seamless Editing}

Seamless editing, particularly in the domains of photo and texture editing, has been an area of significant research focus in the field of computer graphics and digital image processing. The aim is to ensure that modifications or blends in the images or textures are smooth and undetectable, thus maintaining their visual coherence.
A noteworthy work on this task is the ``Poisson Image Editing'' by \citet{Prez2003PoissonIE}, where they showcased a gradient-domain color adjustment technique for managing color inconsistencies in image compositing, thereby ensuring seamless transitions. This work was closely followed by \citet{Agarwala2004InteractiveDP}, which integrates gradient-domain blending with graph cuts, enabling the seamless combination of different sources at interactive speeds for a broad range of compositing applications. Their framework demonstrated success in stitching unrelated photos that possess roughly similar overlapping regions \cite{Kaneva2010CreatingAE}.
Meanwhile, \citet{Kwatra2005TextureOF} introduces ``Texture Optimization'' that allows the seamless transfer of photographic textures from an example to a target image, effectively cloning texture patterns. \citet{Barnes2009PatchMatchAR} put forth ``PatchMatch'', an algorithm that enables seamless reshuffling of image regions. The advent of deep learning led to methods like ``Deep Image Analogy'' by \cite{Liao2017VisualAT}, which uses convolutional neural networks (CNNs) to find semantically meaningful dense correspondences between two input images, providing more advanced solutions for seamless editing.
%
Regarding seamless editing in 3D objects, \citet{rocchini1999multiple} and \citet{dessein2014seamless} propose methods for stitching and blending textures on 3D objects. However, the appearance of NeRFs is encoded in neural networks rather than textures, resulting in challenges to using texture-based approaches. \citet{yu2004mesh} adopts the Poisson equation to implicitly modify the original mesh geometry through gradient field manipulation. While this work offers smooth merging in geometry, the appearance blending is not considered. The application of gradient-domain methods is also extended to the spatio-temporal domain by \citet{WANG200757}, which primarily focuses on video frames.

In essence, traditional editing techniques degenerate when applied directly to NeRFs as image melding algorithms do not account for occlusion and view consistency in 3D space and existing 3D melding approaches lack essential designs for neural implicit fields.


\section{Seamless Neural Radiance Fields}
\label{sec:method}

Our approach towards creating SeamlessNeRF begins with reconstructing multiple NeRFs from individual sets of multi-view images. We choose TensoRF-VM \cite{chen2022tensorf} as the NeRF backbone. After the acquisition of 3D assets represented in NeRFs, we aim to merge them into a single harmonized one.

\subsection{Transforming and Merging Radiance Fields}

The first stage involves the transformation and alignment of distinct NeRFs. This stage receives a user-input transformation which will be applied to join the source field $F_1$ and the target fields $\{F_i\}_{i=2}^N$.
We assume the coordinate system of the merged object is unified with the source field in a global coordinate system. We apply $3\times 4$ affine transformations to align the target fields into the unified global space. This manual alignment plays a role in maintaining the geometric coherence of the merged object, ensuring there are reasonable overlapping areas between the source field and target fields for appearance blending.

Once the transformation is complete, the next stage is to render these aligned fields as a single field $F_{M}: (\mathbf{x}, \mathbf{d}) \rightarrow \left (\mathbf{c}, \sigma\right)$. Since each field's coordinate-to-radiance mapping is independent of each other, a unified mapping for the merged object should be established. Thus, we define the unified field $F_{M}$ as a piece-wise function over multiple sub-fields. Intuitively, this piece-wise function should hold the property that sparser space in the source field should be dominated by the densest field in order to allow the target fields to occupy the empty space in the source field. To achieve this, we define a field selector function based on the density:
\begin{equation}
	S(\mathbf{x}) = \argmax_{i=1..N} {\beta_i\sigma_i(\boldsymbol{M}_i \mathbf{x})},
\end{equation}
where $N$ is the number of the merging fields, $\boldsymbol{M}_i \in \mathbb{R}^{3\times 4}$ is an affine matrix transforming the i-th field to the unified global space (note that the transformation of the source field is an identity matrix), $\beta_i$ is a tunable weight for scaling the precedence order for different fields. Higher $\beta_i$ indicates higher precedence of the i-th field over others even though its original density is lower. Given the selector $S(\mathbf{x})$, the piece-wise function of the unified field $F_{M}: (\mathbf{x}, \mathbf{d}) \rightarrow \left(\mathbf{c}^M_S, \sigma^M_S \right)$ can be written as:
\begin{align}
	\sigma^M_S(\mathbf{x}) &= \sigma_{S(\mathbf{x})}\left(\boldsymbol{M}_{S(\mathbf{x})} \mathbf{x} \right), \\
	\mathbf{c}^M_S(\mathbf{x},\mathbf{d}) &= \mathbf{c}_{S(\mathbf{x})}\left(\boldsymbol{M}_{S(\mathbf{x})} \mathbf{x}, \boldsymbol{M}_{S(\mathbf{x})} \mathbf{d} \right).
\end{align}
It is noteworthy to highlight that both $\mathbf{x}$ and $\mathbf{d}$ are denoted by homogeneous coordinates. The last element of $\mathbf{x}$ is 1 while the last element of $\mathbf{d}$ is 0. With this formulation, the merged radiance field $F_{M}(\mathbf{x}, \mathbf{d}) $ can be rendered with volumetric rendering \cite{max1995optical}:
\begin{align}
	& \mathbf{\hat{C}}(\mathbf{r}(t)) = \sum^K_{j=1} \tau_j \mathbf{c}^M_S(\mathbf{x}_j,\mathbf{d}_j), \\
        & \tau_j = \exp\bigg(-\sum_{i=1}^{j-1}\sigma^M_S(\mathbf{x}_j) \delta_i \bigg) \bigg(1-\exp\big(-\sigma^M_S(\mathbf{x}_j) \delta_j \big)\bigg),
	\label{eq:vol_rendering}
\end{align}
where the rendered color $\mathbf{\hat{C}}(\mathbf{r}(t))$ estimates the image pixel corresponding to the ray $\mathbf{r}(t)=\mathbf{o} + t\mathbf{d}$, $K$ is the number of samples along $\mathbf{r}(t)$, $\delta_j$ is the step length of the $j$-th sample, and $\tau_j$ can be seen as the probability that the ray can reach the $j$-th sample.

\subsection{Evaluating Radiance via Closest Ray Sampling}
\label{sec:sampling}

The direct merge of multiple NeRFs is followed by an appearance optimization over the target radiance fields. For each coordinate $\mathbf{x}$, rendering its radiance necessitates a view direction $\mathbf{d}$. If $\mathbf{x}$ is sampled from a ray originating from a camera, the process of obtaining the corresponding $\mathbf{d}$ is straightforward. However, if the goal is to optimize the radiance across the entire 3D space without assuming a specific viewing camera, the view direction $\mathbf{d}$ is agnostic.

In order to address this challenge, we propose a simple yet effective \textit{Closest Ray Sampling} solution that hinges on sampling the view direction $\mathbf{d}$ from the entire spectrum of the training dataset. This scheme can match the distribution of view directions with the one seen in the multi-view captures. Specifically, for each point $\mathbf{x}$, we determine its view direction as the orientation of the closest training camera ray. Suppose $\left\{ \mathbf{r}_j\left(\mathbf{p}_j, \mathbf{\widetilde{d}}_j\right) \right\}_{j=1}^R$ is the set of all camera rays in the training set. The Euclidean distance $\|D_{\mathbf{r}_j,\mathbf{x}}\|_2$ between a ray $\mathbf{r}_j\left(\mathbf{p}_j, \mathbf{\widetilde{d}}_j\right)$ and a sample point $\mathbf{x}$ is defined as:
\begin{equation}
	p^{e}_j = \mathbf{p}_j-\left[(\mathbf{p}_j-\mathbf{x}) \cdot \mathbf{\widetilde{d}}_j\right] \mathbf{\widetilde{d}}_j, ~\text{and}~
	D_{\mathbf{r}_j,\mathbf{x}} = p^{e}_j - \mathbf{x},
	\label{eq:ray_distance}
\end{equation}
where $p^{e}_j$ is the nearest point from the ray $\mathbf{r}_j$ to the sample point $\mathbf{x}$, $D_{\mathbf{r},q}\in\mathbf{R}^{3}$ is the offset vector from $\mathbf{x}$ to $p^{e}$. The designated view direction for $\mathbf{x}$ is given as $\mathbf{\widetilde{d}}_j\ast$, where $j\ast = \argmin_j {\|D_{\mathbf{r}_j,\mathbf{x}}\|_2}$.


\begin{figure}[t]
	\centering
	\includegraphics[width=0.95\linewidth]{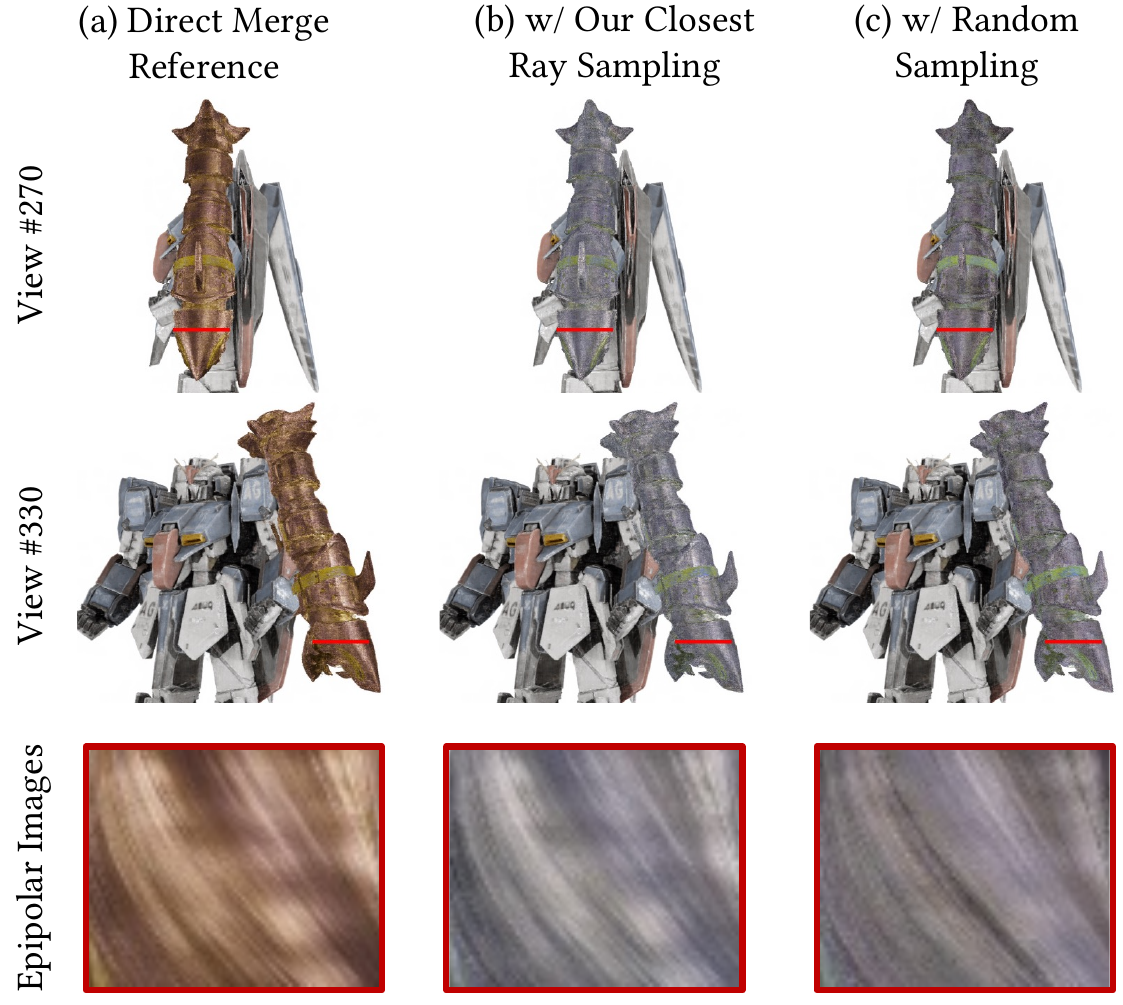}
	\caption{View-dependent effects on the blended radiance fields via different view-direction sampling schemes.}
	\label{fig:viewdependent-effects}
\end{figure}

\paragraph{Impacts on View-dependent Effects} View directions are adopted to encode the view-dependent effects, e.g., specular highlights. Therefore, a feasible scheme for sampling view directions is obligated to maintain the view-dependent effects on the original object. Fig. \ref{fig:viewdependent-effects} presents a case of shiny object appearance blending. In addition to our proposed scheme, we also provide a straightforward baseline where view directions are randomly sampled. Comparing the reference views (a), our scheme (b) is capable of preserving the original view-dependent specular effects while the baseline scheme (c) smoothens the specular highlights. We also plot (pseudo) epipolar images \cite{verbin2022ref,bemana2022eikonal} of the red scanline segment. As shown by the last row of Fig. \ref{fig:viewdependent-effects}, the patterns on the epipolar images obtained through our scheme are more similar to the reference one, in contrast to the baseline scheme.

\begin{figure*}[t!]
	\centering
	\includegraphics[width=\textwidth]{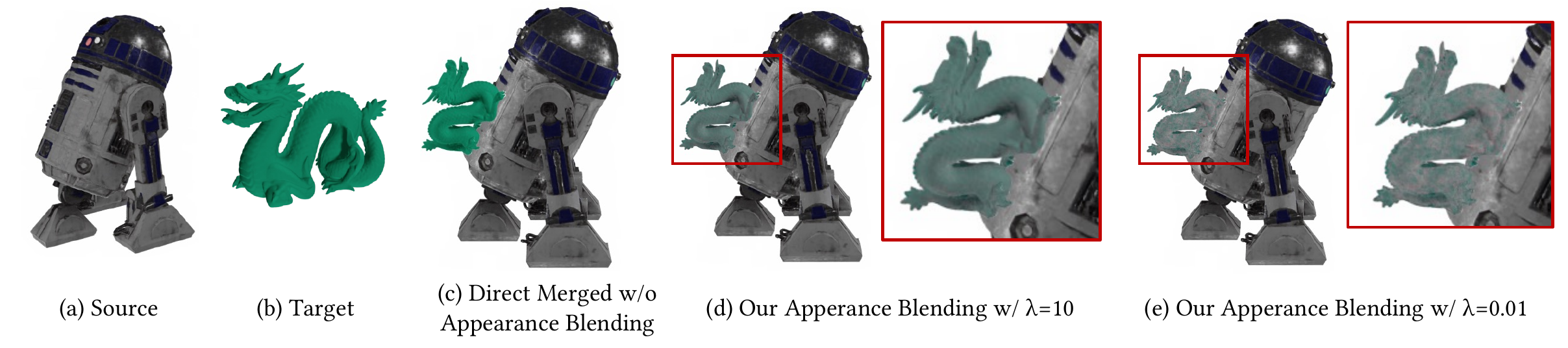}
	\caption{Ablation study on the weights of gradient loss. The first two columns (a) and (b) display the source and target objects. The column (c) gives a reference view of direct merge without blending. In (d), we show the blending results under $\lambda=10$. In (e), we show the blending results when decreasing $\lambda$ to $0.01$.}
	\label{fig:ablation_loss_diff}
\end{figure*}

\begin{figure}[t]
	\centering
	\begin{subfigure}{.24\linewidth}
		\centering
		\includegraphics[trim={120 80 50 80},clip,width=\textwidth]{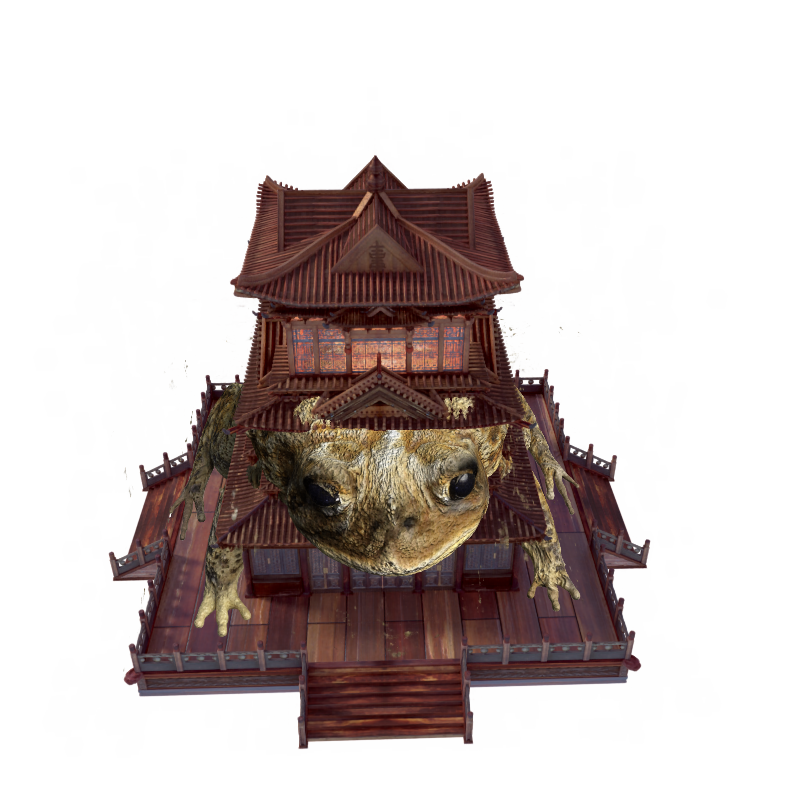}
		\includegraphics[trim={120 80 50 80},clip,width=\textwidth]{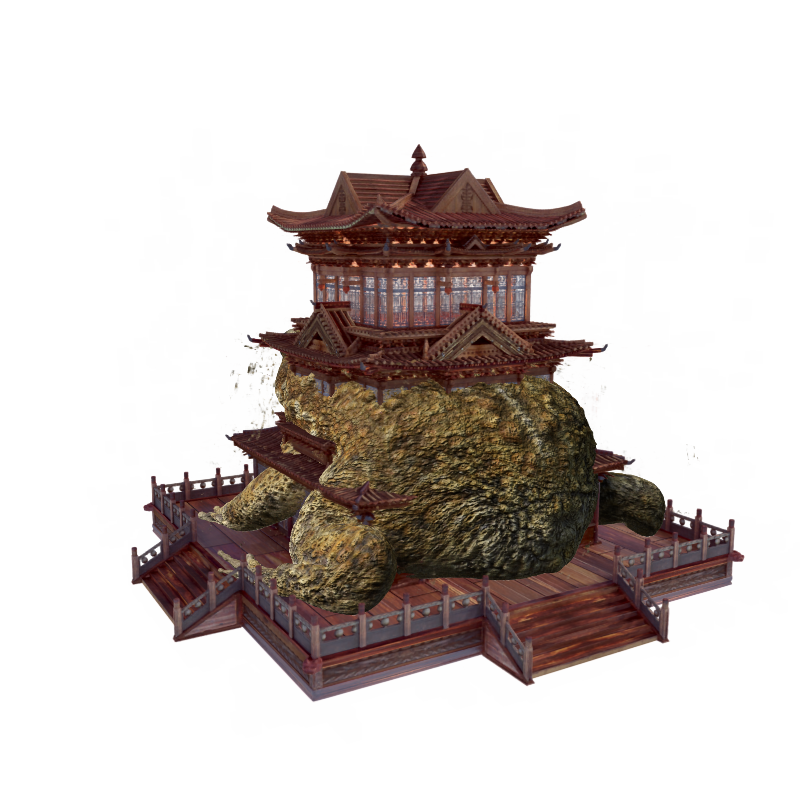}
		\caption{Direct Merge}
	\end{subfigure}
	\begin{subfigure}{.24\linewidth}
		\centering
		\includegraphics[trim={120 80 50 80},clip,width=\textwidth]{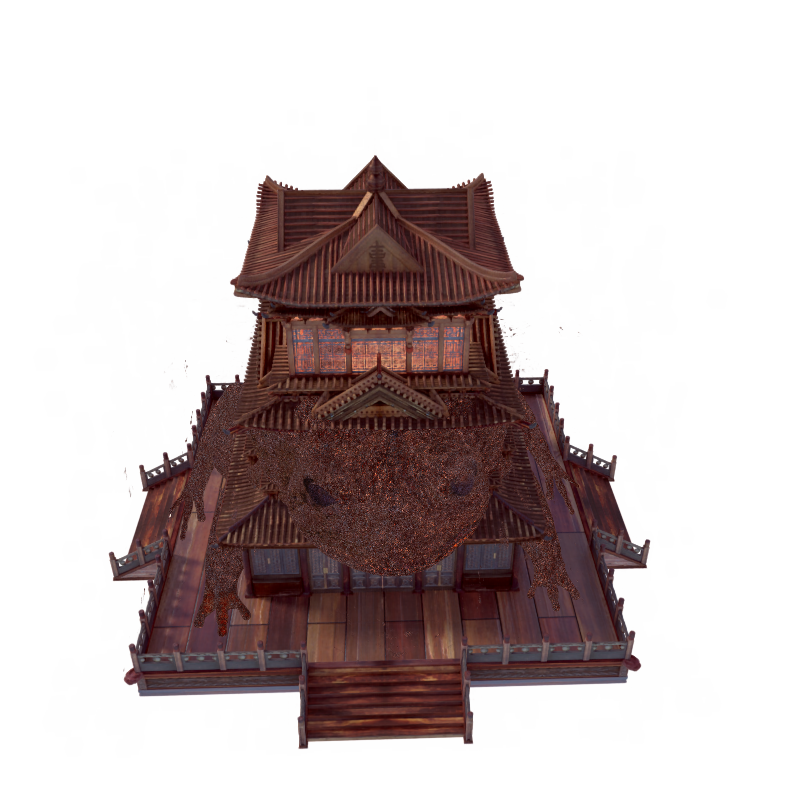}
		\includegraphics[trim={120 80 50 80},clip,width=\textwidth]{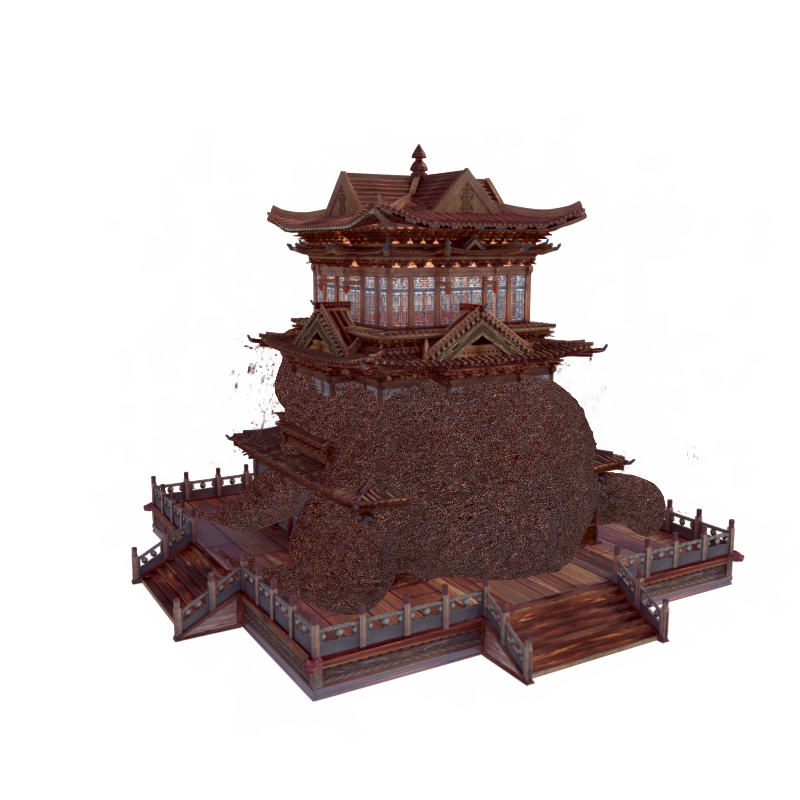}
		\caption{w/o $\mathcal{L}_{grad}$}
		\label{fig:ab1-figure:nograd}
	\end{subfigure}
	\begin{subfigure}{.24\linewidth}
		\centering
		\includegraphics[trim={120 80 50 80},clip,width=\textwidth]{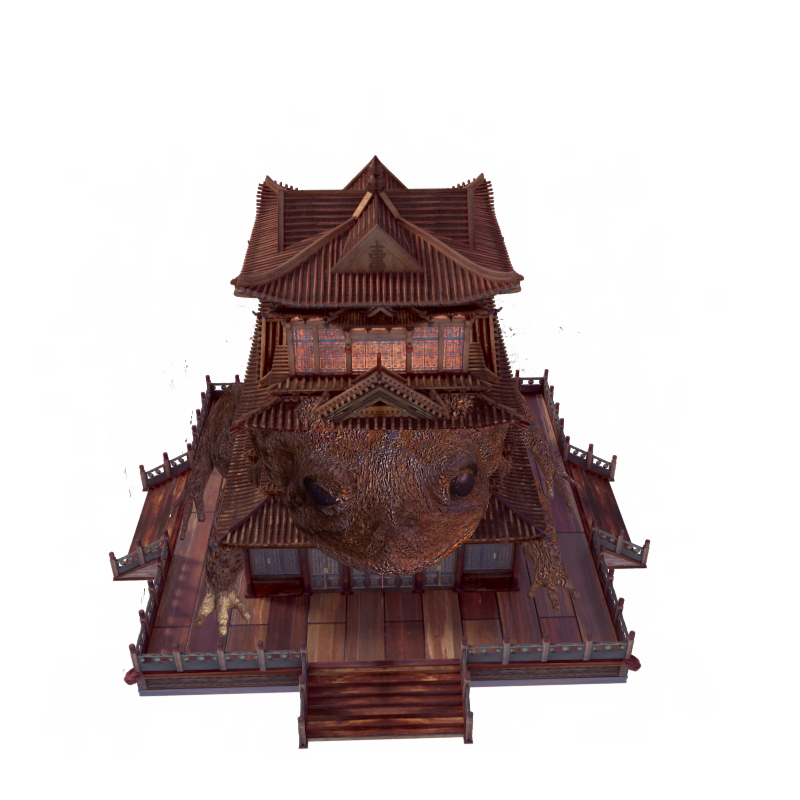}
		\includegraphics[trim={120 80 50 80},clip,width=\textwidth]{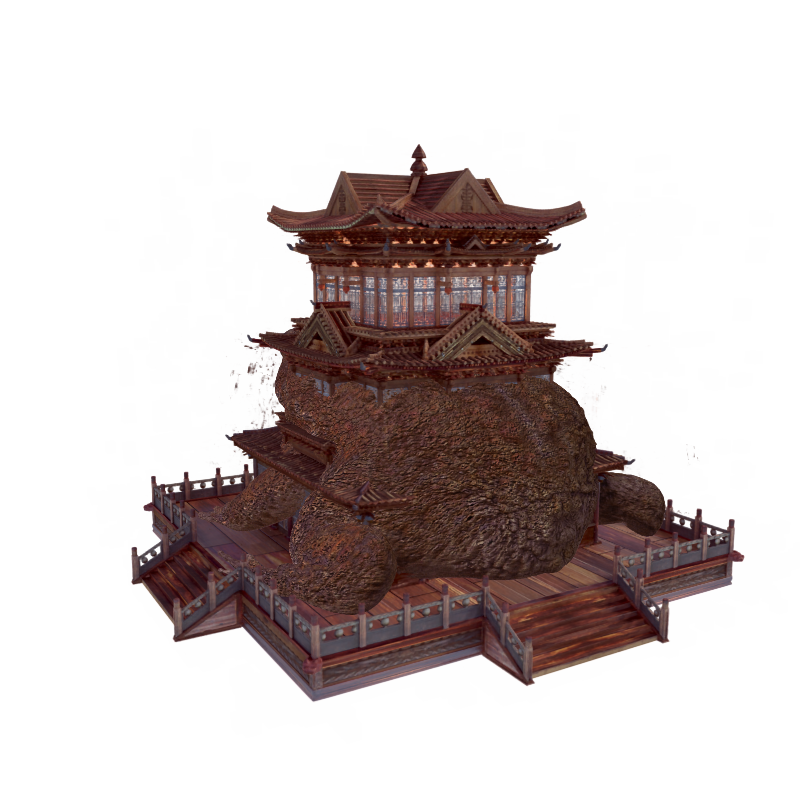}
		\caption{w/o $\mathcal{L}_{color}$}
        \label{fig:ab1-figure:nocolor}
	\end{subfigure}
	\begin{subfigure}{.24\linewidth}
		\centering
		\includegraphics[trim={120 80 50 80},clip,width=\textwidth]{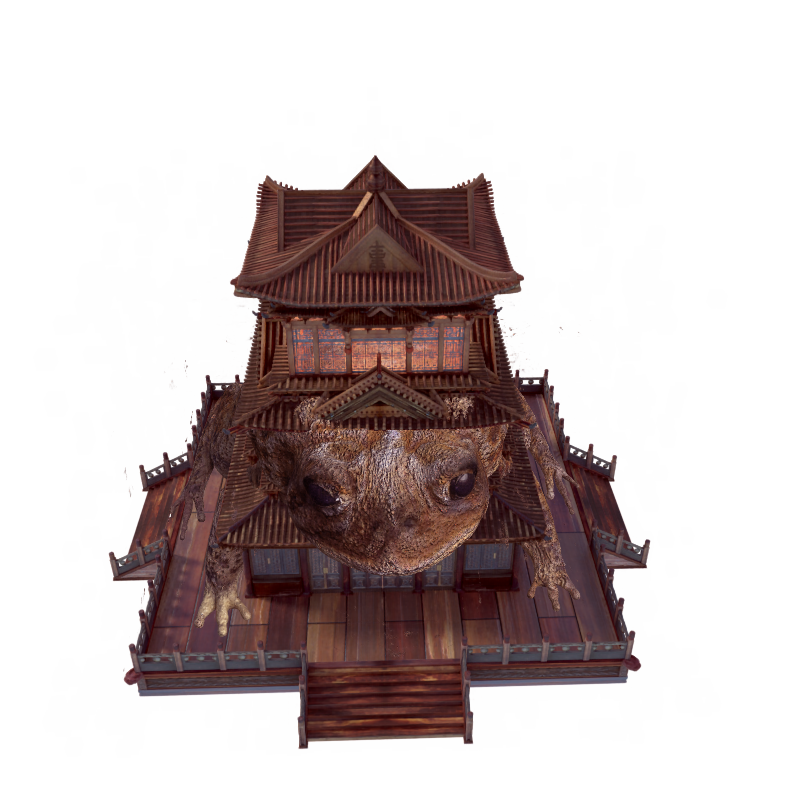}
		\includegraphics[trim={120 80 50 80},clip,width=\textwidth]{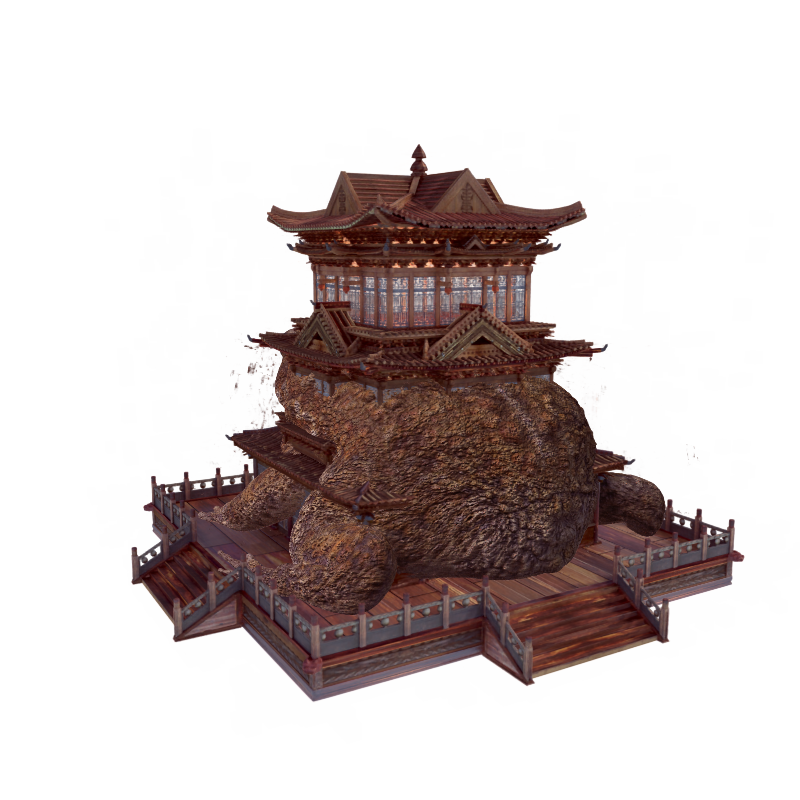}
		\caption{Full Model}
        \label{fig:ab1-figure:fullmodel}
	\end{subfigure}
	\caption{Ablation study on loss functions. The column (a) gives two reference views of direct merge. The column (b) shows results when disabling the gradient loss. The column (c) shows results without optimizing the color loss. The column (d) shows results when enabling both the gradient loss and the color loss.}
	\label{fig:ab1-figure}
\end{figure}

\subsection{Boundary Detection and Radiance Pinning}

The key to generating seamless NeRFs iss to harmonize the color of the target fields with the color of the source field at the boundaries.
We draw inspiration from the concept of contact areas to define the boundary regions. Conceptually, these are areas where the density of the radiance field is simultaneously dominated by the source field and the target fields. Formally, we define the boundary regions $\{\partial B_i\}_{i=2}^N$ as follows:
\begin{equation}
	\partial B_i  \leftarrow \left\{\mathbf{x} \mid S(x)=1 ~\text{and}~ \sigma_i(\boldsymbol{M}_i \mathbf{x}) > T_{\text{th}} \right \},
\end{equation}
where the threshold $T_{\text{th}}$ signifies the minimal density at which a point in the field is still considered non-empty. This usage is akin to the threshold application in the marching cubes algorithm \cite{lorensen1998marching}, a technique employed for isosurface extraction.

These boundaries $\partial B_i$ are typically locations where the transition between the source and target radiance fields occurs, and hence, can be potential sites with visual inconsistency when these fields are merged directly.
To address this, we introduce a strategy termed \textit{Boundary Radiance Pinning}. The primary objective of this strategy is to minimize the discrepancies between the source field and target fields.
Boundary Radiance Pinning functions by enforcing an agreement in radiance color among all target fields and the source field within the boundary region.
Specifically, we minimize the loss between the target radiance color and the source radiance color $\mathbf{c}_1(\mathbf{x},\mathbf{d})$ within the boundary:
\begin{equation}
	\begin{split}
		\mathcal{L}_{color} &= \sum_{\mathbf{x} \in \partial B_i} \left \| \mathbf{c}^M_i(\mathbf{x},\mathbf{d}) - \mathbf{c}_1(\mathbf{x},\mathbf{d}) \right \|_2^2.
	\end{split}
	\label{eq:nerf_obj_func}
\end{equation}
The view direction $\mathbf{d}$ is obtained by our proposed scheme described in Sec. \ref{sec:sampling}. By ensuring alignment in radiance color, appearance continuity on the joined regions is maintained.

\subsection{Gradient Propagation of Radiance}

The boundary radiance pinning will disturb the original details on target fields (see Fig. \ref{fig:ab1-figure:nograd}). In order to preserve the rich textures on the target field, we further add an optimization objective to propagate radiance through the gradient field. The gradients can represent the variation in radiance across different spatial locations. This is particularly critical as many of the patterns and textures are implied in the gradient field of the radiance, as opposed to a single color value.
To do this, we first evaluate the gradients of the radiance field within the target regions. This is performed by differentiating the target radiance field w.r.t. the spatial coordinates $\mathbf{x}$. After obtaining the gradients of the original target field, we use these gradients as the optimization target for the merged field in the unified global space. The loss function to regularize the gradients of the merged field is formulated as:
%
\vspace{5px}
\begin{equation}
	\begin{split}
		\mathcal{L}_{grad} &= \sum_{\mathbf{x},~~S(\mathbf{x}) > 1} \left \| \nabla_{\mathbf{x}} \mathbf{c}^M_S(\mathbf{x},\mathbf{d}) - \nabla_{\mathbf{x}} \mathbf{c}_i(\boldsymbol{M}_i\mathbf{x},\boldsymbol{M}_i\mathbf{d}) \right \|_F^2,
	\end{split}
	\label{eq:nerf_grad_func}
        \vspace{5px}
\end{equation}
where the corresponding view direction $\mathbf{d}$ can be obtained via the sampling scheme described in Sec. \ref{sec:sampling}, $i=S(\mathbf{x})$ signifies the field index at $\mathbf{x}$, and $\left\|\cdot\right\|_F$ is the Frobenius norm. In practice, this involves an iterative optimization procedure, where we aim to fine-tune the target field such that its gradients align with the gradients of the source field at the boundary region and simultaneously maintain the gradient of its original appearance at other regions.
However, taking the analytic second-order derivative poses a challenge when using gradient-based optimizers to solve objectives involving gradients. Furthermore, the gradients w.r.t. spatial coordinates $\mathbf{x}$ are derived from tri-linear interpolation in feature grids, which are not spatially continuous \cite{li2023neuralangelo}.
In this regard, we opt for a discrete solution.  Denote superscript $k = 1, 2, 3$ as the index of XYZ dimensions. We can calculate the finite difference in radiance color at the position $\mathbf{x}$ by:
\vspace{5px}
\begin{equation}
    \mathcal{V}_{\mathbf{x}^k} = \mathbf{c}_i(\boldsymbol{M}_i \mathbf{x},\boldsymbol{M}_i \mathbf{d}) - \mathbf{c}_i\left(\boldsymbol{M}_i (\mathbf{x}+\Delta_k),\boldsymbol{M}_i \mathbf{d}\right).
    \vspace{5px}
\end{equation}

\begin{figure*}[t!]
	\centering
	\includegraphics[width=\textwidth]{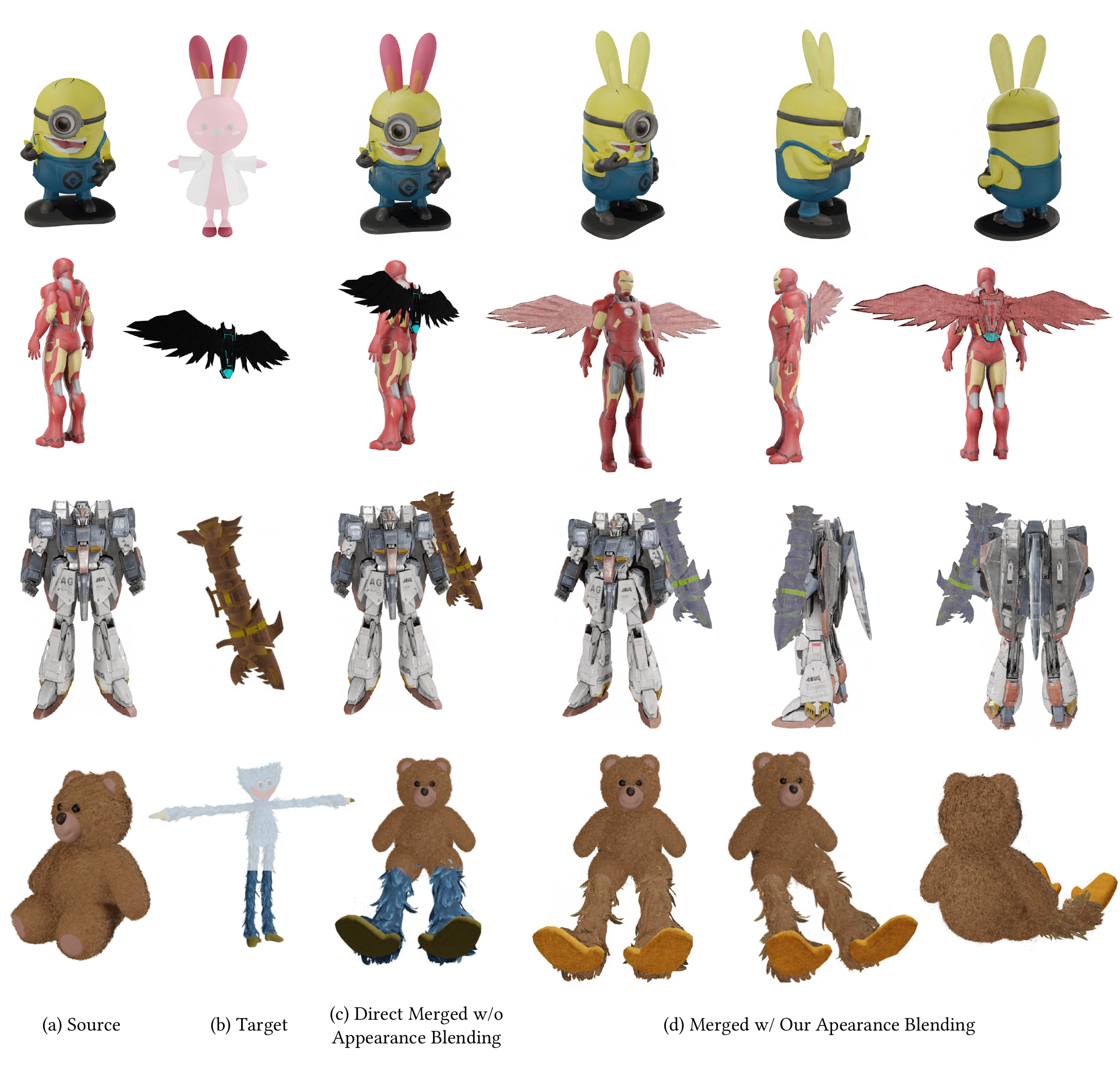}
	\caption{Showcase of our apperance blending results. In (a) and (b), we present reference images of source and target radiance fields, respectively. The column (c) shows results of direct merge of the source and target. The last three columns in (d) exhibit three views of the merged and appearance blended radiance fields.}
	\label{fig:morevis-figure}
\end{figure*}
Here, $\mathbf{x}+\Delta_k$ is a neighborhood point of $\mathbf{x}$ along the k-th dimension. Then we obtain the discrete version of gradient loss:
\begin{equation}
	\begin{split}
		\mathcal{L}_{grad} &= \sum_{\mathbf{x},~~S(\mathbf{x}) > 1} \sum_{k=1}^3 \left \| \mathbf{c}^M_S(\mathbf{x},\mathbf{d}) -  \mathbf{c}^M_S(\mathbf{x}+\Delta_k,\mathbf{d}) - \mathcal{V}_{\mathbf{x}^k} \right \|_2^2.
	\end{split}
	\label{eq:nerf_grad_dis}
\end{equation}
In the above equation, the difference between $\mathbf{c}^M_S(\mathbf{x},\mathbf{d})$ and $\mathbf{c}^M_S(\mathbf{x}+\Delta_k,\mathbf{d})$ approximates the radiance gradient of the merged field at the position $\mathbf{x}$ along the k-th dimension.
The sum is taken over all sample points in the merged field and XYZ dimensions.

By minimizing this gradient loss, the textures and patterns implied in the gradient field can be preserved. Meanwhile, the color scheme from the source field can be propagated to the target fields, resulting in a seamless transition of radiance across the boundary.

\subsection{Optimization and Network Architecture}


To establish the final objective function, we combine the gradient loss with the color loss. The overall loss of our optimization process is formulated as a weighted sum, defined as follows:
\begin{equation}
	\mathcal{L} = \mathcal{L}_{color} + \lambda \mathcal{L}_{grad}.
\end{equation}
In this equation, the parameter $\lambda$ is utilized to balance the relative contribution of color loss and gradient loss to the overall objective. In our experiments, we set the value of $\lambda$ as 0.1.

To conduct the appearance optimization of the target fields,  we introduce a ControlNet-style \cite{zhang2023adding} architecture to facilitate the fine-tuning process. This involves adopting the original network parameters as the initialization of a trainable copy and appending a ``zero convolution'' to connect the trainable copy to the original networks as a render branch. During the appearance blending process, the parameters of the original networks will be frozen, serving as an initial model and evaluator of the original gradient field. The ControlNet-style network branch is introduced to learn the incremental modifications, which are added to the original appearance via the ``zero convolution''. The choice of ``zero convolution'' stems from its ability to retain the original output and progressively fuse modifications to the appearance at the beginning of fine-tuning.
Fig. \ref{fig:pipeline-figure} presents the entire pipeline of our method.

\begin{figure}[t]
	\centering
	\includegraphics[trim={0 48 00 42},width=0.88\linewidth]{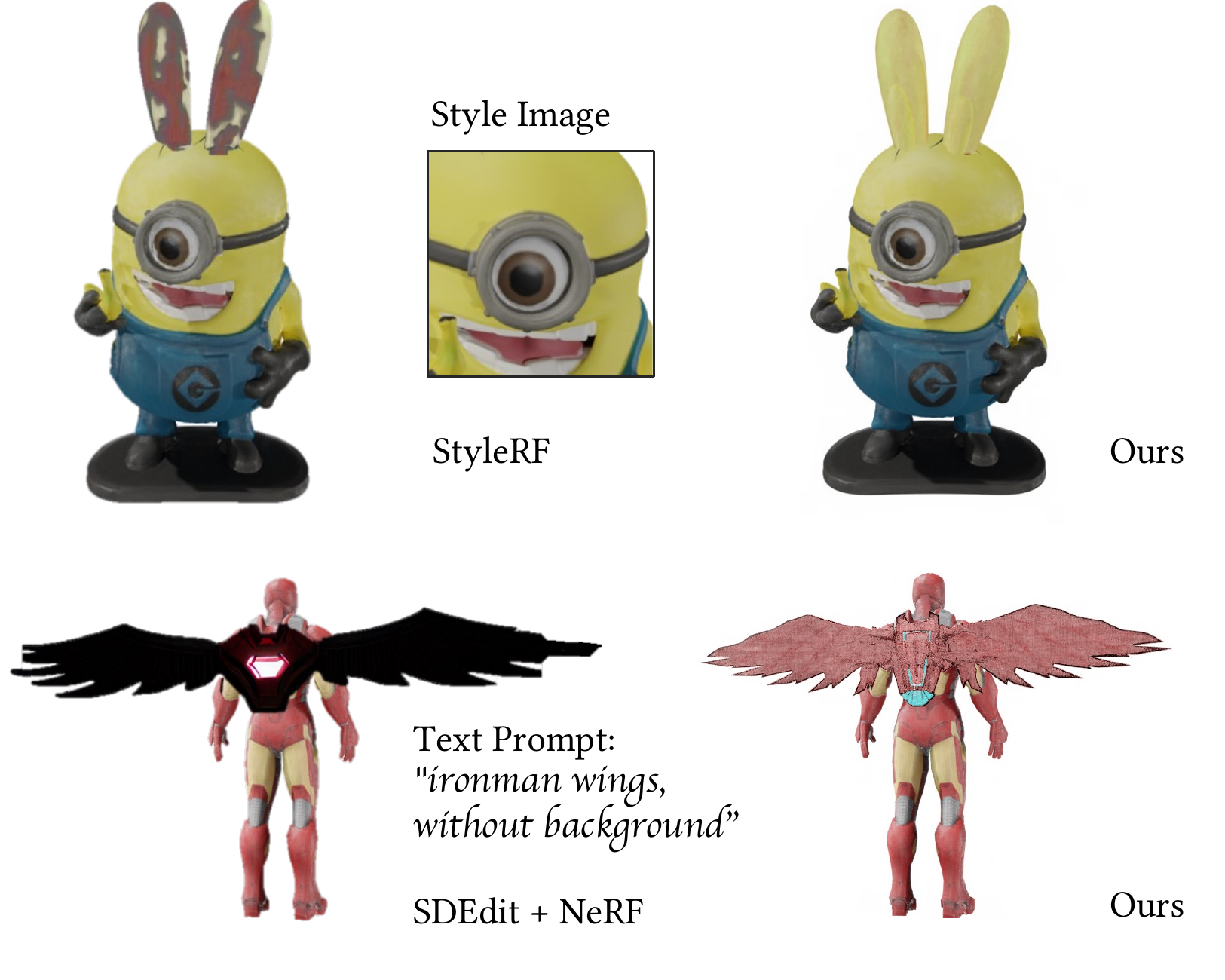}
	\caption{Comparisons between our approach and baseline methods. For the first case, the baseline StyleRF \cite{liu2023stylerf} stylizes the target radiance fields with the source one. For the second case, we adopt SDEdit \cite{meng2021sdedit} to perform text-guided image-to-image generation of target radiance field renderings.}
	\label{fig:cmp_baselines}
\end{figure}

\section{Experiments}
\label{sec:experiments}

To evaluate the performance of our proposed approach, our experiments are conducted on multiple 3D models from the Objaverse \cite{deitke2023objaverse} dataset. Our implementation is built on the backbone of TensoRF. Nevertheless, since our approach operates directly on the output of radiance fields, it does not depend on specific radiance field modeling. Results of our approach with Na\"ive NeRF \cite{mildenhall2021nerf}, Instant-NGP \cite{muller2022instant}, and DirectVoxGO \cite{sun2022direct} backbones are present in Fig. \ref{fig:ft_backbones}.

\begin{figure}[t]
	\centering
	\includegraphics[width=0.95\linewidth]{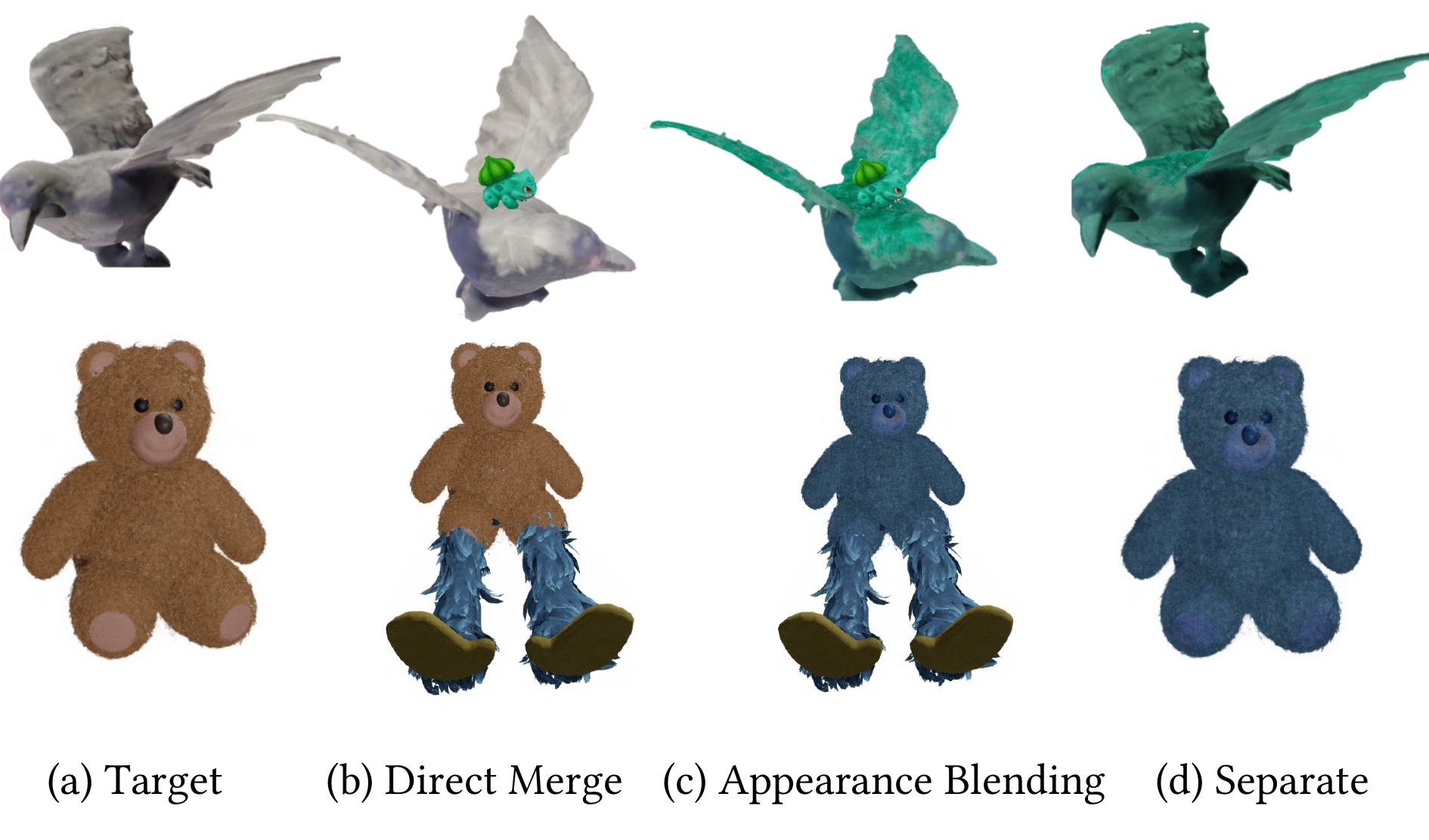}
	\caption{Color scheme cloning results via our appearance blending approach.}
	\label{fig:app_color_clone}
\end{figure}

\begin{figure}[t]
	\centering
	\begin{subfigure}{.32\linewidth}
		\centering
		\includegraphics[trim={320 180 180 400},clip,width=\textwidth]{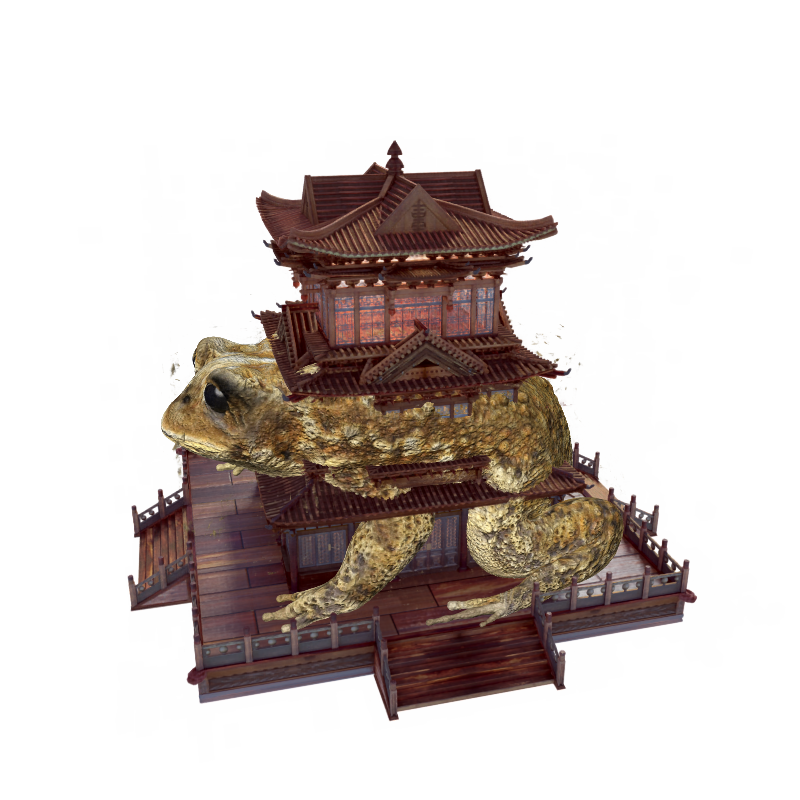}
		\caption{w/o App. Blending}
	\end{subfigure}
	\begin{subfigure}{.32\linewidth}
		\centering
		\includegraphics[trim={320 180 180 400},clip,width=\textwidth]{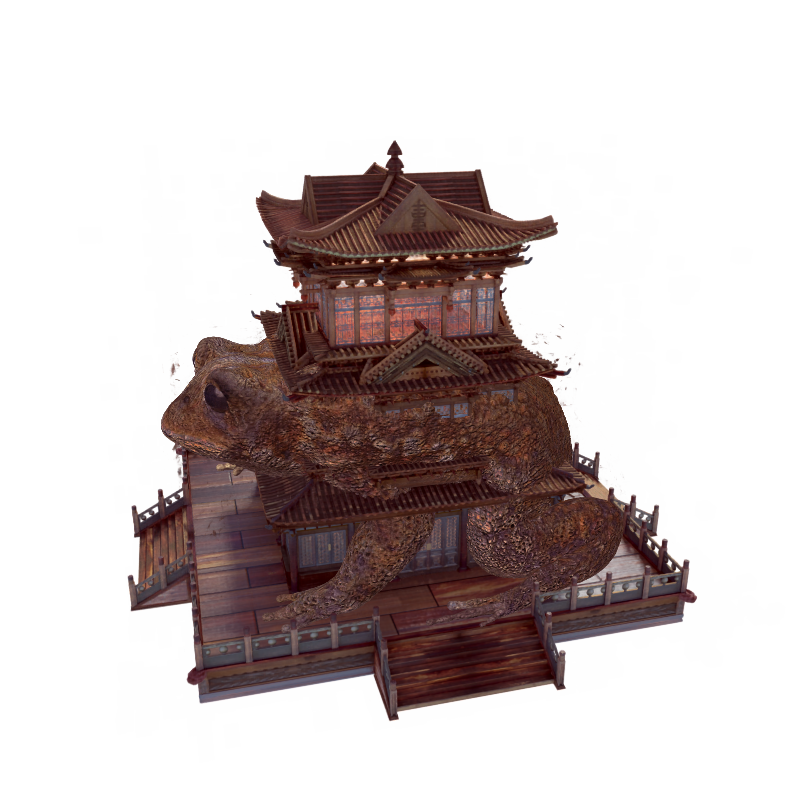}
		\caption{Opt. MLPs Only}
	\end{subfigure}
	\begin{subfigure}{.32\linewidth}
		\centering
		\includegraphics[trim={320 180 180 400},clip,width=\textwidth]{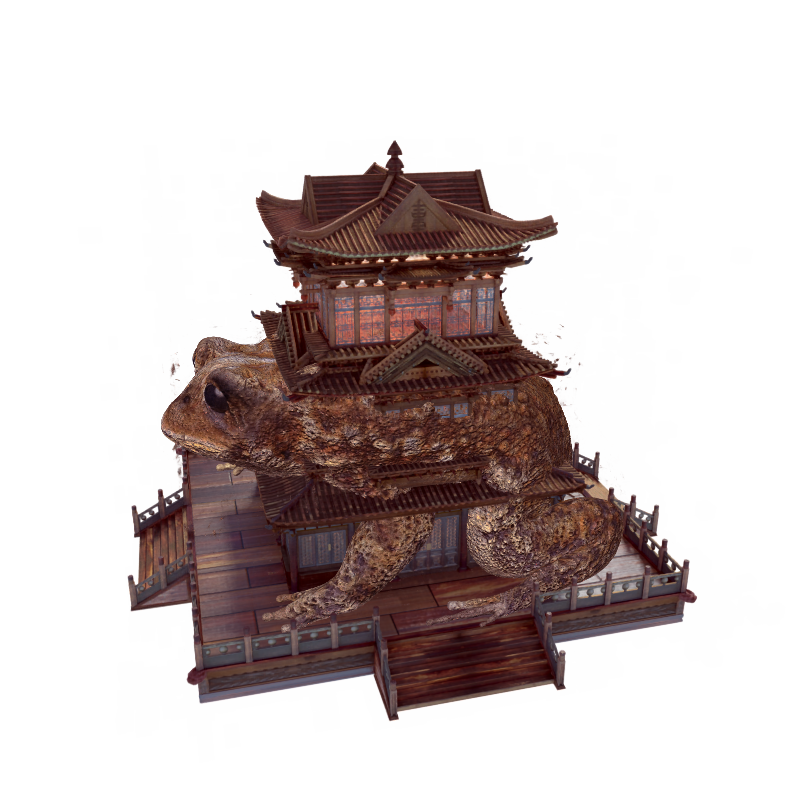}
		\caption{Opt. MLPs + VM}
	\end{subfigure}
	\caption{Ablation study on different configurations of the trainable copy.}
	\label{fig:ab2-figure}
\end{figure}

\subsection{Qualitative Comparison}

Since there is no existing method of seamless appearance blending in the context of radiance fields, we build two baselines with recent emerging techniques to validate the effectiveness of our approach. The first baseline attempts to perform 3D style transfer from the source to target. We train StyleRF \cite{liu2023stylerf} for zero-shot NeRF stylization. For the second baseline, we utilize the diffusion model to conduct image-to-image generation \cite{meng2021sdedit} guided by an editing text prompt and rendered images. Fig. \ref{fig:cmp_baselines} shows two comparison cases. In the first case, StyleRF fails to seamlessly transfer the color scheme from the ``Minion'' to the ``rabbit ear'' since it is not capable of extracting style information from an ordinary image. In the second case, we use a text prompt to generate a view based on the original rendering in order to match our editing goal. We can see the generated ``wing'' contains elements of ``Ironman'' but still fails to yield a seamless merge.

\subsection{Ablation Study}

\paragraph{Gradient Loss for Radiance Propagation}
%
As demonstrated in Fig. \ref{fig:ab1-figure:nograd}, upon removing the gradient loss function, we observed a remarkable deterioration in the quality of radiance propagation, resulting in artifacts on the texture. After adding the gradient loss, the small details on the texture implied in the gradient field are well preserved. Moreover, Fig. \ref{fig:ablation_loss_diff} shows the impacts of the gradient loss weight. When imposing a strong constraint ($\lambda = 10$), the radiance gradients are maintained while the impacts from the source object are weakened. When decreasing the weight to $0.01$, the appearance from the source object dominates the appearance blending and violates the original radiance gradients.
%

\paragraph{Boundary Radiance Pinning}
In this study, we opted to omit the color loss component in our objective to analyze the impacts of boundary radiance pinning. Fig. \ref{fig:ab1-figure:nocolor} displays the result without the color loss. Since boundary pinning is disabled, the target radiance field will be optimized through mixed gradients from the source field, which harms the rich lighting and textual details in the original target field. Thus, combining color loss and gradient loss can achieve the best blending performance.
\paragraph{Configurations of Trainable Copies}
We also ablate different trainable copy configurations. Our TensoRF-VM \cite{chen2022tensorf} backbone consists of VM (vector-matrices) factors and shading MLPs, where VM factors compute the feature of each query point and shading MLPs render the feature into RGB color. Fig. \ref{fig:ab2-figure} shows that the blending performance will compromise when freezing the VM factors and only optimizing the shading module. Thus, a trainable VM copy is essential for high-quality blending.





\subsection{Visual Results and Applications}

We showcase the results of our SeamlessNeRF on multiple cases, as exhibited in Fig. \ref{fig:morevis-figure}.
We can see that the our approach can generate seamless part NeRF stitching on various challenging cases. We also include the visualizations of the optimization process in Fig. \ref{fig:ft_process}.

\paragraph{Applications}
Our proposed approach demonstrates the potential for expansion into 3D creation and editing workflows, offering advantages in the productivity of 3D modeling. For instance, consider the scenario where artists intend to restore the arms of the ``Venus de Milo'' sculpture, as shown in Fig. \ref{fig:teaser}. In such a case, our approach enables the seamless blending of pre-existing arm assets onto the sculpture, expediting the modeling process compared to starting from scratch. As a more practical case, the ``Toad Temple'' in Fig. \ref{fig:ab1-figure:fullmodel} exhibits a building model stitched from two existing 3D assets, which can be employed as a game element. Users can also customize cartoon characters (the first case in Fig. \ref{fig:morevis-figure}) without time-consuming modeling and re-texturing. In addition, Fig. \ref{fig:app_color_clone} illustrates a color scheme transferring application of our approach, where the target objects clone the appearance of the source objects. 


\section{Conclusion}
\label{sec:conclusions}

This paper presents SeamlessNeRF, a novel approach to seamless stitching of multiple NeRFs into a unified 3D scene. By extending the principles of gradient-domain blending to NeRFs, our method can propagate appearance from the source field to the target field. Our experiments validate the effectiveness of our approach in achieving seamless merges across diverse objects. The successful application of Poisson Editing to NeRFs represents a significant advancement in the realm of 3D scene composition and editing.

\paragraph{Limitations and Future Work}
Since our current work doesn't explicitly factor in lighting, it may struggle to handle the cases when the lighting and material properties of the source and target fields are inconsistently matched. To disentangle the impacts of lighting, we can apply our propagation to the albedo component with a NeRF backbone supporting decomposed lighting and materials, which is a promising topic for future work.

%
%
%
%

\begin{acks}
This work was supported in part by Science, Technology and Innovation Commission of Shenzhen Municipality Project No. SGDX202205\\30111201008, in part by Hong Kong Research Grants Council Project No. T45-401/22-N, in part by NSFC-62172348, in part by Outstanding Young Fund of Guangdong Province with No. 2023B1515020055, in part by Shenzhen General Project with No. JCYJ20220530143604010, and in part by CCF-Tencent Open Research Fund. 
This work used and adapted 3D models created and shared by 
\href{https://sketchfab.com/3d-models/female-sculpture-29a8029d6022411ca4a6911ca13f9b3b}{Coen.Fransen (CC BY)},
\href{https://sketchfab.com/3d-models/female-marble-statue-polished-but-old-41b024813afd49fb827d942219d89b93}{Nikolai J\'{o}nasson (CC BY)},
\href{https://sketchfab.com/3d-models/rabbit-squat-08d4bf632019457bbfb87c3a9b3b9803}{hanjum (CC BY)},
\href{https://sketchfab.com/3d-models/iron-man-mark7-b8d5a0156ef04cb7800d19f6bfd9842b}{mwilson1 (CC BY)},
\href{https://sketchfab.com/3d-models/bulbasaur-64815cda802746b8b1be2e2246db4b35}{fongoose (CC BY-NC)},
\href{https://sketchfab.com/3d-models/r2d2-7719d93e76ca40a5a8774f5530a7aecf}{jesuskrisis (CC BY)},
\href{https://sketchfab.com/3d-models/huggy-wuggy-normal-with-animations-d5d4bdb1e1ca4c2dbed988f90310c368}{Edward Johnson 3 (CC BY)},
\href{https://sketchfab.com/3d-models/msz-006-zeta-gundam-c79a648bdd1b4ccfbf11f2dc5f836a66}{cosmos28 (CC BY)},
\href{https://sketchfab.com/3d-models/teddy-bear-a635e23e0e2f4f0992fbf595bce223ef }{firmasyahh (CC BY)},
\href{https://sketchfab.com/3d-models/wing-d627c48b676741d08144255295368816}{SamkyClance (CC BY)},
\href{https://sketchfab.com/3d-models/42fe263b572e4402bba90a41a3110a3d}{Hristo (CC BY)},
\href{https://sketchfab.com/3d-models/d6b85e8dc4b54269b3df6c7e1e5541ba}{Maurice Svay (CC BY)},
\href{https://sketchfab.com/3d-models/minion-dcf09f31b3174c91997f748abb1f0018}{racush (CC BY)},
and \href{https://sketchfab.com/3d-models/blackbird-af1034a3b3094eb2a6654d2c6c9d87bc}{David Wigforss (CC BY)}.
\end{acks}

\balance
\bibliographystyle{ACM-Reference-Format}
\bibliography{sample-bibliography}

\begin{figure*}[t]
	\centering
	\includegraphics[width=\textwidth]{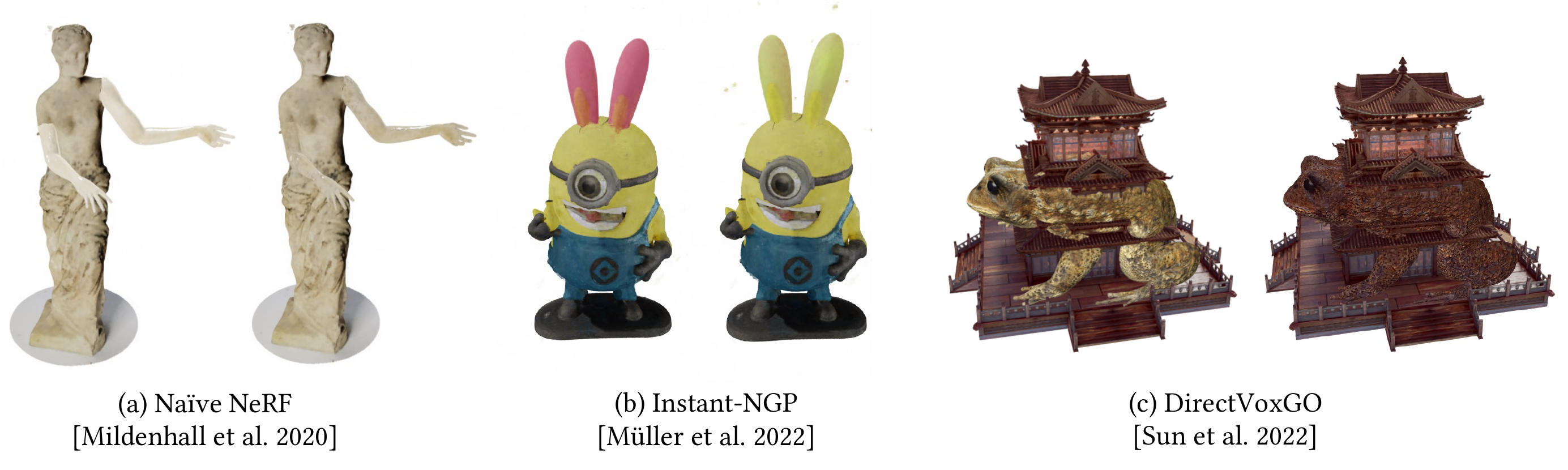}
	\caption{Appearance blending results with the three backbones. Our method can be implemented across various radiance field backbones. Although NeRF variants differ in radiance field parameterization, they all encode scenes as a mapping from coordinates to densities and RGB values via differentiable neural encodings. As our approach operates directly on the output, specific parameterizations do not impact our gradient-based blending framework. In addition to TensoRF which has been already used in our main experiments, we test our approach on Na\"ive NeRF \cite{mildenhall2021nerf}, Instant-NGP \cite{muller2022instant}, and DirectVoxGO \cite{sun2022direct}.}
	\label{fig:ft_backbones}
\end{figure*}

\begin{figure*}[t]
	\centering
	\includegraphics[width=\textwidth]{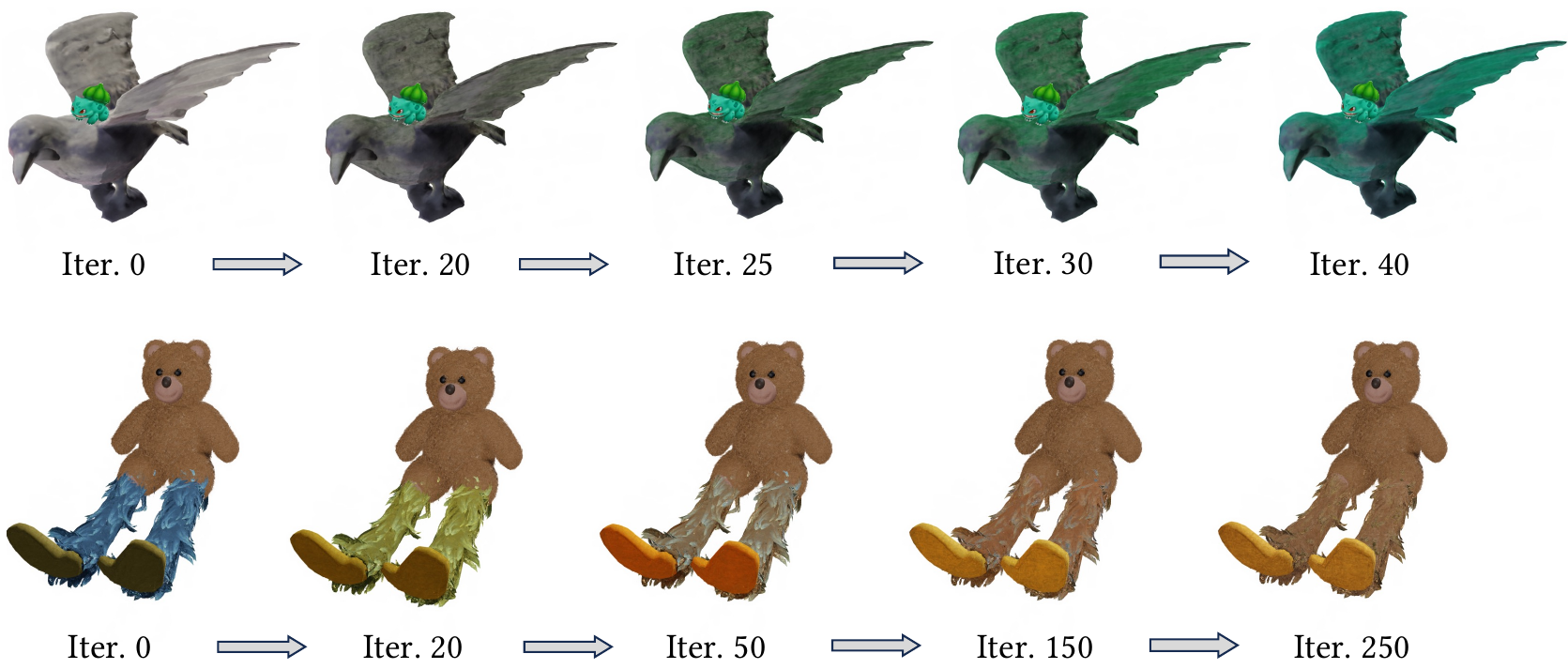}
	\caption{Visualizes the fine-tuning processes of two representative cases. The fine-tuning of the simpler case (the first row) will converge within 50 iterations. While the case with more complex textures (the second row) consumes more fine-tuning time to simultaneously optimize the gradient loss and color loss.}
	\label{fig:ft_process}
\end{figure*}

\end{document}